\documentclass[conference]{IEEEtran}
\IEEEoverridecommandlockouts
\usepackage{cite}
\usepackage{amsmath,amssymb,amsfonts}
\usepackage{graphicx}
\usepackage{textcomp}
\usepackage{xcolor}
\usepackage{booktabs}
\usepackage{subcaption}
\usepackage{comment}
\usepackage{makecell}
\usepackage{url}
\usepackage{hyperref}

\def\BibTeX{{\rm B\kern-.05em{\sc i\kern-.025em b}\kern-.08em
T\kern-.1667em\lower.7ex\hbox{E}\kern-.125emX}}

\begin{document}

\title{Component-Aware Differential Privacy for Federated Multilingual Speech-LLMs}

    \author{
\IEEEauthorblockN{1\textsuperscript{st} Jordi Luque}
\IEEEauthorblockA{\textit{Scientific Research} \\ \textit{Telefónica Innovación Digital} \\
Barcelona, Spain \\
jordi.luque@telefonica.com}
\and
\IEEEauthorblockN{2\textsuperscript{nd} Fernando López}
\IEEEauthorblockA{\textit{Telefónica Innovación Digital}\\
\textit{Universidad Autónoma de Madrid}\\
Madrid, Spain \\
fernando.lopez@telefonica.com}
\and
\IEEEauthorblockN{3\textsuperscript{rd} Aleix Sant}
\IEEEauthorblockA{\textit{Scientific Research}\\ \textit{Telefónica Innovación Digital}\\
\textit{Universitat Politècnica de Catalunya}\\
Barcelona, Spain \\
aleix.santsavall@telefonica.com}

}

\maketitle

\begin{abstract}
Per-layer differential privacy (DP) clipping improves gradient fidelity in federated learning by allocating per-matrix clipping budgets proportional to parameter count.  We show that this recipe breaks for speech large language
models (speech-LLMs), when the acoustic encoder and the language decoder differ by an order of magnitude in update norm.  Single-pool per-layer methods suffer \emph{cross-component budget collapse}, dragging word error rate (WER) far from flat global clipping or collapsing training entirely. When the norm imbalance is milder, adaptive single-pool methods partially recover, confirming that collapse severity scales with the inter-component norm ratio.  We empirically diagnose the root cause across six per-layer methods and three speech-LLM architectures. We then propose \emph{$\alpha$-split}, a two-pool allocation that normalises encoder and LLM parameters into independent pools, and show that joint $\ell_2$ sensitivity and the original $(\varepsilon,\delta)$-DP guarantee are unchanged. At architecture-calibrated $\alpha$, our method recovers WER utility compared to flat DP, while granting the encoder $4.47{\times}$ tighter per-component noise protection against speaker voice-based gradient-inversion attacks at only $+2.6\%$ LLM noise overhead.
\end{abstract}


\section{Introduction}
\label{sec:intro}
Large language models (LLMs) and speech foundation models are increasingly combined to build end-to-end speech-LLM systems for automatic speech recognition (ASR), spoken understanding, and conversational speech applications~\cite{cui-etal-2025-recent, 10.1609/aaai.v40i36.40318}. These architectures typically couple three heterogeneous components: an acoustic encoder, a cross-modal connector, and a language decoder~\cite{DBLP:journals/corr/abs-2402-08846, DBLP:conf/iclr/ZhangZLZQ24}. While this modular design improves transferability and downstream performance, it also introduces optimization asymmetries across components, especially in distributed and privacy-sensitive training settings.

Federated learning (FL)~\cite{mcmahan2017communication} is a natural paradigm for speech applications because raw audio is privacy-critical, bandwidth-heavy, and often constrained by data governance policies. However, practical FL deployments for speech-LLMs face two challenges. First, client data are strongly non-IID across speakers, accents, microphones, and acoustic environments, which induces unstable update distributions~\cite{kairouz2021advances}. Second, differential privacy (DP) mechanisms~\cite{10.1007/11787006_1,dwork2014algorithmic,mcmahan2018learning}, particularly clipping-based methods, can alter optimization dynamics in ways that are not yet well understood for multimodal, multi-component models.

To achieve formal $(\varepsilon, \delta)$-DP~\cite{dwork2014algorithmic} in federated learning, \emph{gradient clipping} serves as a fundamental mathematical requirement: before the server injects calibrated Gaussian noise, it must strictly bound the $\ell_2$ sensitivity of each client update to a clipping budget $C$. Standard DP-FL applies a single \emph{global} $C$ to the full concatenated update. While most prior FL+DP studies report aggregate utility-privacy trade-offs using global hyperparameters~\cite{kairouz2021advances}, this approach is architecturally oblivious---encoder LoRA adapters, connector projections, and LLM adapters are clipped identically despite having fundamentally different update magnitudes and sensitivity requirements.
Recent work~\cite{DBLP:conf/nips/PelikanAFSTBL25} demonstrated that \emph{per-layer} clipping, where each parameter matrix receives an individual budget proportional to its size, substantially outperforms flat global clipping for homogeneous single-component ASR. We show this recipe breaks for multimodal speech-LLMs.

We demonstrate that this state-of-the-art recipe~\cite{DBLP:conf/nips/PelikanAFSTBL25} systematically \emph{breaks down} in modular Speech-LLMs when the encoder/LLM update norm ratio is large enough. In such situations, the encoder's vast number of parameters structurally dilutes the LLM's clipping budget. We investigate six single-pool per-layer methods, reporting that all of them suffer severe cross-component budget collapse, driving the utility Word Error Rate (WER) either far from the flat global clipping ceiling or collapsing training entirely. To resolve this bottleneck, we formalize the necessity of component-aware privacy allocation. Our key contributions are:
\begin{enumerate}
\item \textbf{A precise diagnosis of \emph{cross-component budget collapse}}: We empirically demonstrate that when encoder/LLM norm imbalances are extreme ($\ge 10\times$), single-pool methods suffer severe budget collapse during critical early warmup rounds, necessitating a structural solution.
\item \textbf{An adaptive single-pool formulation}: We adapt the size-proportional per-layer clipping concept~\cite{DBLP:conf/nips/PelikanAFSTBL25} to parameterised LoRA matrices and introduce a dynamic gradient norms tracking mechanism. We show that this adaptive formulation dynamically corrects mild encoder/LLM norm imbalances in Speech-LLMs.
\item \textbf{The structural \emph{$\alpha$-split} DP design}: We propose a component-aware strategy that decouples gradient normalisation into two independent pools, allocating budgets of $C\sqrt{\alpha}$ to the acoustic encoder and $C\sqrt{1-\alpha}$ to the LLM decoder. We prove that this separation entirely prevents budget collapse while strictly preserving the joint $\ell_2$ sensitivity and the global $(\varepsilon,\delta)$-DP mathematical guarantees.
\item \textbf{Delineation of operating regimes and asymmetric biometric privacy}: We map out clear deployment guidelines: Adaptive layer-based is optimal for mild imbalances (measured $\approx\!1.7{\times}$ for Voxtral-Mini-3B), while $\alpha$-split is strictly required for extreme imbalances (measured $\approx\!12{\times}$ for Whisper+TinyLlama/EuroLLM). Furthermore, we show that the $\alpha$-split formulation uniquely grants the acoustic encoder tighter effective noise multiplier, achieving stronger acoustic protection for sensitive biometric data at a negligible noise overhead to the LLM.
\end{enumerate}

\section{Speech-LLM Framework and Private Federated Setup}
\label{sec:background}

\subsection{Speech-LLM Architecture}
\label{sec:arch}

\emph{Speech-LLMs} couple three heterogeneous components: an acoustic encoder
($\mathcal{E}$), a cross-modal connector ($\mathcal{C}$), and a language
decoder ($\mathcal{L}$), producing a transcript as:

\begin{equation}
  \hat{\mathbf{y}} = \mathcal{L}\left([\mathcal{C}(\mathcal{E}(\mathbf{x}));\,\mathbf{E}_{\mathrm{text}}]\right).
  \label{eq:model}
\end{equation}

\begin{figure}[t]
  \centering
  \includegraphics[width=1\linewidth]{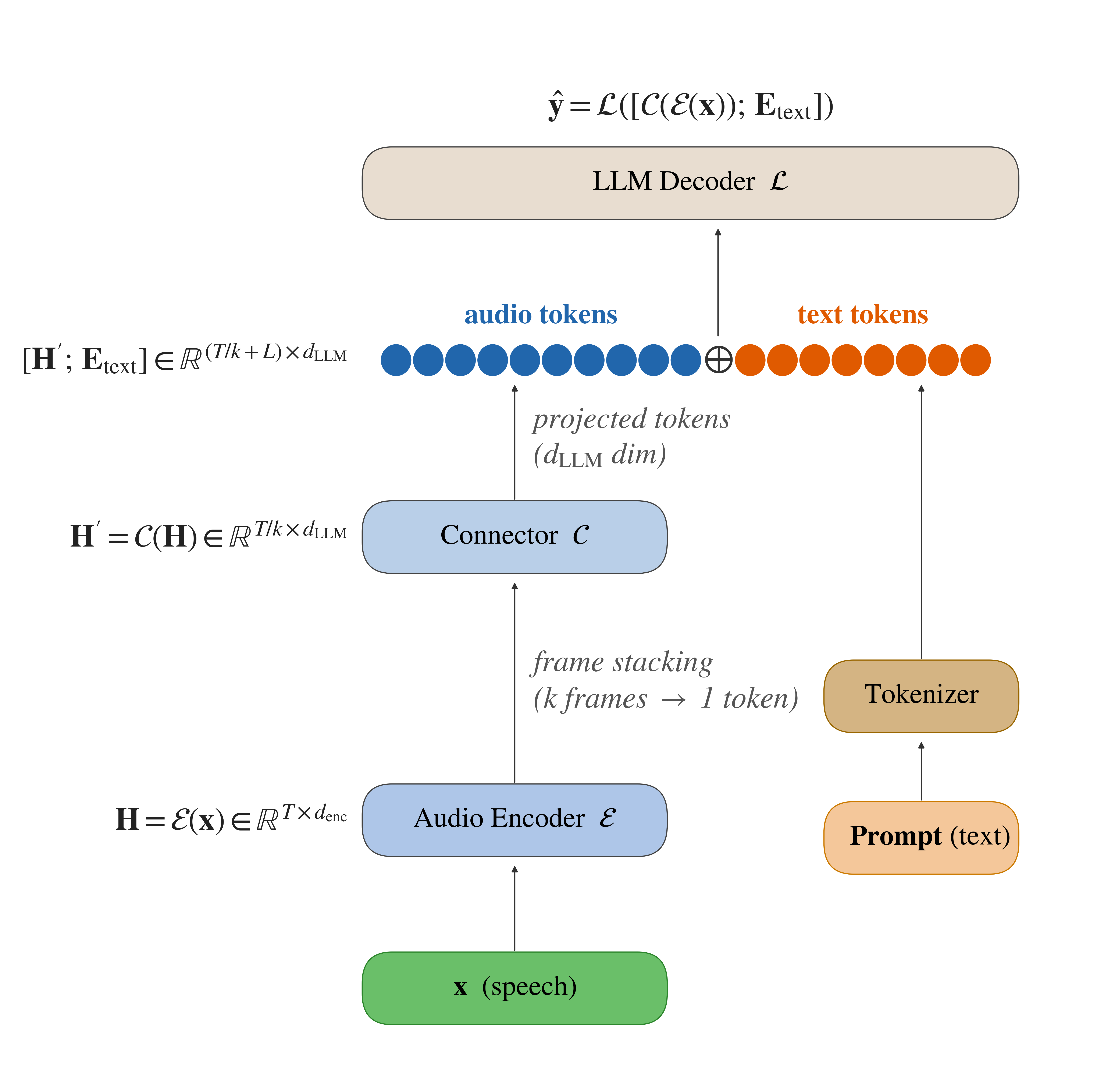}
  \caption{Speech-LLM architecture. The audio encoder $\mathcal{E}$ maps raw speech $\mathbf{x}$ to frame-level representations $\mathbf{H}$. The connector $\mathcal{C}$ downsamples frames, reducing sequence length by $k$. The LLM decoder $\mathcal{L}$ generates the transcript        $\hat{\mathbf{y}}$ conditioned on the projected tokens and a task prompt.}
  \label{fig:arch}
\end{figure}

Figure~\ref{fig:arch} summarises the full pipeline. All three components are trained jointly by minimising the standard
autoregressive cross-entropy loss over the ground-truth transcript tokens
$\mathbf{y} = (y_1, \ldots, y_S)$:
\begin{equation}
  \mathcal{L}_{\mathrm{CE}}(\boldsymbol{\theta}) =
  -\sum_{s=1}^{S} \log p_{\boldsymbol{\theta}}\left(y_s \mid
    \mathcal{C}(\mathcal{E}(\mathbf{x})),\,
    \mathbf{E}_{\mathrm{text}},\,
    y_{<s}\right),
  \label{eq:loss}
\end{equation}
where $\boldsymbol{\theta}$ collects all trainable parameters (LoRA
adapters and connector projection), and $y_{<s}$ denotes the preceding
tokens supplied via teacher forcing.  The loss is computed over
transcript tokens; audio tokens and prompt embeddings appear as
conditioning context. We study a three-component \emph{Speech-LLMs}: a Whisper large-v3-turbo
encoder~\cite{radford2023whisper} $\mathcal{E}$, a linear connector
$\mathcal{C}$, and a TinyLlama-1.1B~\cite{zhang2024tinyllama} (or
EuroLLM-1.7B-Instruct~\cite{martins2024eurollm}) decoder $\mathcal{L}$; and additionally evaluate \emph{Voxtral-Mini-3B}~\cite{mistral2025voxtral} an end-to-end multimodal Speech-LLM, composed of a Whisper-large-v3-based audio encoder and a Ministral-3B~\cite{liu2026ministral3} LLM, a 30-layer Llama-style text decoder; jointly pretrained on audio understanding and ASR tasks.  Voxtral uses a different connector $\mathcal{C}$, downsampling the audio by a factor of 4, followed by a 2-layer MLP projector; a deeper, higher-compression connector. 

\subsection{Central DP-FL Trust Model}
\label{sec:dp-background}

We adopt the \emph{central DP} trust model~\cite{dwork2014algorithmic,
mcmahan2018learning}, i.e. the aggregation server is trusted, each client clips its own update to bound $\ell_2$ sensitivity, and the server adds calibrated Gaussian noise to the aggregate, thus privacy is guaranteed against any external observer of the released model sequence. Each client clips its update before transmission:
\begin{equation}
  \widetilde{\Delta\boldsymbol{\theta}}^{(i)} =
  \Delta\boldsymbol{\theta}^{(i)} \cdot
  \min\!\left(1,\;\frac{C}{\left\|\Delta\boldsymbol{\theta}^{(i)}\right\|_2}\right),
  \label{eq:global_clip}
\end{equation}
and the server aggregates with Gaussian noise:
\begin{equation}
  \boldsymbol{\theta}^{(t+1)} =
  \boldsymbol{\theta}^{(t)} +
  \frac{1}{n}\sum_{i=1}^{n}\widetilde{\Delta\boldsymbol{\theta}}^{(i)}
  + \mathcal{N}\!\left(\mathbf{0},\,\frac{\sigma^2 C^2}{n^2}\,\mathbf{I}\right).
  \label{eq:gaussian_mech}
\end{equation}
The $\ell_2$ sensitivity of the clipped aggregate is $C/n$, so
$\sigma$ is the noise multiplier relative to the sensitivity.  All
experiments use $C{=}1.0$, $\sigma{=}0.1$; privacy accounting uses
Rényi DP~\cite{mironov2017renyi}. Note that in the case of FedAvg~\cite{mcmahan2017communication} for Speech-LLMs, each client $i$ computes the
full parameter delta as a flat concatenation:
\begin{equation}
  \Delta\boldsymbol{\theta}^{(i)} =
  \bigl[\Delta\boldsymbol{\theta}_\mathcal{E}^{(i)},\;
        \Delta\boldsymbol{\theta}_\mathcal{C}^{(i)},\;
        \Delta\boldsymbol{\theta}_\mathcal{L}^{(i)}\bigr].
  \label{eq:flat_delta}
\end{equation}

\begin{table}[t]
\centering
\caption{LoRA parameter $p$ using $rank{=}8$ for Whisper + TinyLlama Speech-LLM, applied to queries ($q$), keys ($k)$ and values ($v$) in attention heads.}
\label{tab:params}
\setlength{\tabcolsep}{4pt}
\begin{tabular}{lrrr}
\toprule
\textbf{Component} & \textbf{Matrices} & \textbf{Avg.\ params} & \textbf{Total} \\
\midrule
Encoder ($\mathcal{E}$, q/k/v, 32L) & 64 & 10,240 & 1,966,080 \\
LLM ($\mathcal{L}$, q/v, 22L)        & 44 & 16,384 &   720,896 \\
Connector ($\mathcal{C}$, weight+bias)&  2 &  2,048 &     4,096 \\
\midrule
\textbf{Total (H)}                        & 110 & ---   & 2,691,072 \\
\bottomrule
\end{tabular}
\end{table}
Table~\ref{tab:params} summarises the LoRA parameter counts.  The encoder represents 73.1\% of all LoRA parameters, yet each encoder matrix is considerably \emph{smaller} than an LLM attention matrix.

\subsection{Heterogeneous Dataset and Speaker Partitioning}
\label{sec:data}

\subsubsection{Corpus}
All experiments use the Multilingual LibriSpeech (MLS)
corpus~\cite{pratap2020mls}, an audiobook corpus covering 8 European languages derived from LibriVox recordings.  We use the official MLS \texttt{train} splits as the federated training pool (685.7~h total),
the MLS \texttt{dev} split for validation during training, and the MLS \texttt{test} split as the held-out evaluation benchmark (138~h, 19,492 samples). 

\subsubsection{Speaker Data Partition (non-IID)}
\label{sec:b1partition}

We perform a stratified speaker-based partition that assigns all utterances of a single MLS speaker to one client. With $K{=}316$ clients, this creates the strongest possible non-IID distribution: each client's data is drawn from a single acoustic identity, language, and recording environment, producing simultaneous \emph{linguistic} (each client speaks at most one language) and \emph{acoustic} (microphone, room, speaking rate) heterogeneity. Due to the LibriVox origin of MLS, 8 of 316 clients (2.5\%) correspond to speakers also present in the MLS test split, accounting for 4,747 of 169,586 training samples (2.8\%). All FL experiments in this work use this partition.

\subsection{Federated Optimization Configuration}
\label{sec:trainconfig}

All FL experiments use the Flower simulation
framework with Ray as the backend ~\cite{beutel2020flower}.  
All FL experiments train for $T{=}40$ global rounds across a total population of $K = 316$ speaker-partitioned clients; in each round, a random cohort of $n \approx 94$ clients (sampling rate $q = 0.3$). Local clients perform $E{=}10$ local epochs of fine-tuning before aggregation, using AdamW optimiser with maximum learning rate $\eta{=}10^{-4}$, cosine decay, batch size 16. Evaluation uses the MLS test split with overall WER reported across all 8 languages combined. All models are fine-tuned with LoRA~\cite{hu2022lora} adapters applied to all query, key, value for encoder and query, value for LLM decoder heads.
%
%
Unlike the encoders and LLM decoders, which carry rich pretrained representations, we initialise the multimodal connector from scratch, except for Voxtral, and must learn to bridge audio and text modalities entirely from the federated fine-tuning data.

\begin{figure}[!t]
  \centering
  \includegraphics[width=\linewidth]{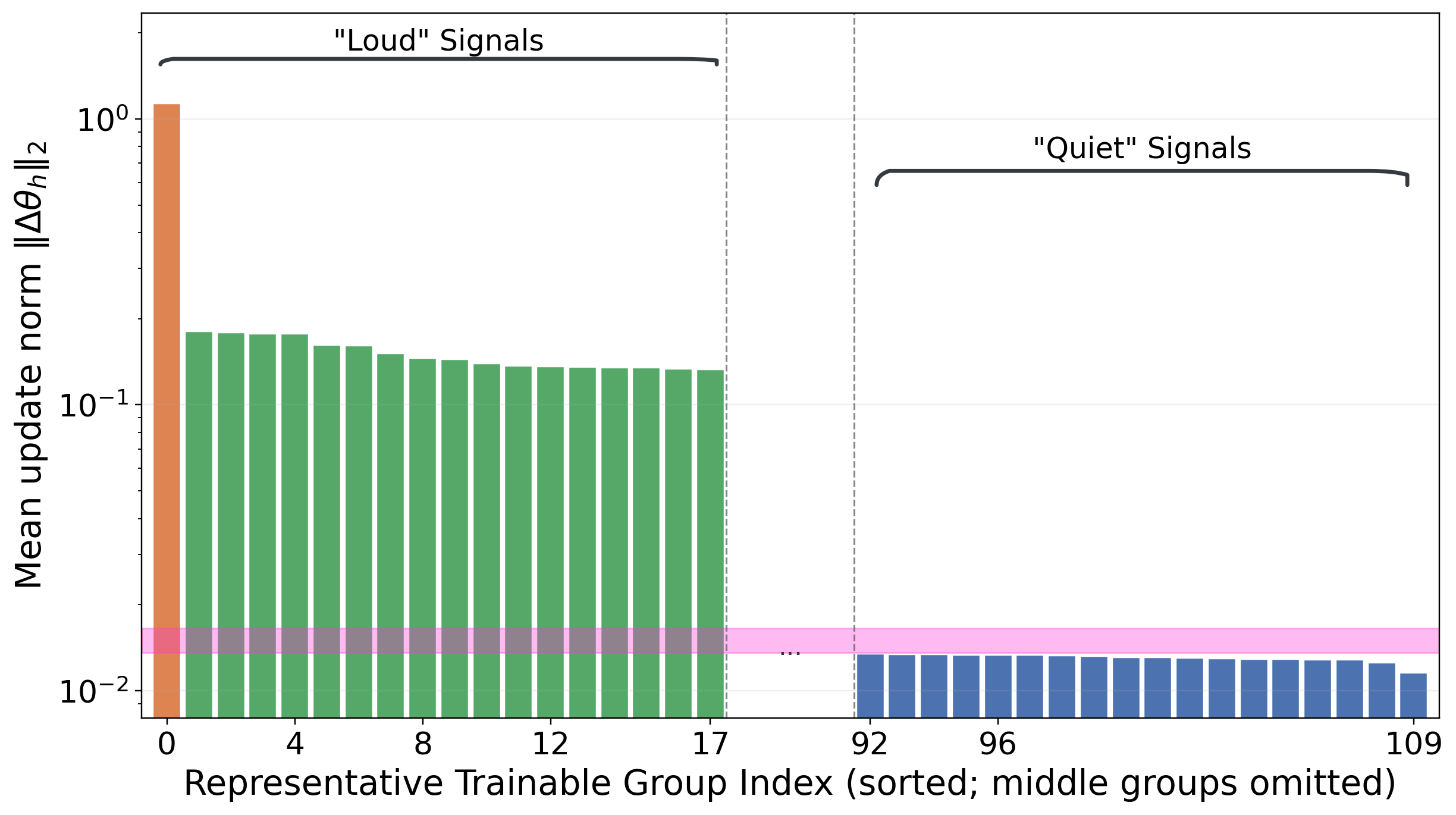}
  \caption{Mean update norm $\|\Delta\theta_h\|_2$ per-Trainable Group $h$, i.e. LoRA matrix (log scale, sorted descending, with no-DP and all components unfrozen) for the Whisper + TinyLlama Speech-LLM. Green bars stand for LLM LoRA matrices. Blue correspond to encoder LoRA matrices. Orange (index 0) to connector (single full-rank matrix). Pink band: Gaussian noise floor $\sigma C/\sqrt{n}{\approx}0.01$. 
  }
  \label{fig:norm-heterogeneity}
\end{figure}

\section{Cross-Component Budget Collapse}
\label{sec:collapse}

\subsection{Empirical Gradient-Norm Asymmetries}
\label{sec:math}

With flat or global clipping, see Eq. \eqref{eq:global_clip}, every LoRA matrix is scaled by the \emph{same} ratio scalar $r$:

\begin{equation}
\widetilde{\Delta\boldsymbol{\theta}}_k^{(i)} = r^{(i)}\Delta\boldsymbol{\theta}_k^{(i)},
\end{equation}
with $r^{(i)}= \min\!\left(1,\,\frac{C}{\|\Delta\boldsymbol{\theta}^{(i)}\|_2}\right)$, the per-client $i$ ratio.

 

Fig.~\ref{fig:norm-heterogeneity} depicts the empirical per-matrix update
norms from a federated training non-DP with all components unfrozen. Two clusters are clearly separated by an order of
magnitude: LLM matrices (green, ${\approx}0.1$--$0.2$) and
encoder LoRA matrices (blue, ${\approx}0.01$--$0.02$).
The pink band marks the Gaussian noise floor
$\sigma C/\sqrt{n} \approx 0.01$: encoder updates already sit at the
noise level under flat DP, while LLM updates remain well above it. When the LLM pool dominates $\|\Delta\boldsymbol{\theta}^{(i)}\|_2$,
$r^{(i)}$ is driven toward $C/\|\Delta\boldsymbol{\theta}_\mathcal{L}^{(i)}\|_2$,
and the encoder pool, with norms already $10\times$ smaller, is
suppressed further toward the noise floor. This ${\sim}10\times$ per-matrix norm gap drives the cross-component budget collapse, where encoder matrices collectively absorb a disproportionate share of the $C^2$ budget.
\subsection{SNR Suppression under Global Clipping}
\label{sec:math2}
\begin{figure*}[t]
  \centering
  \includegraphics[width=\linewidth]{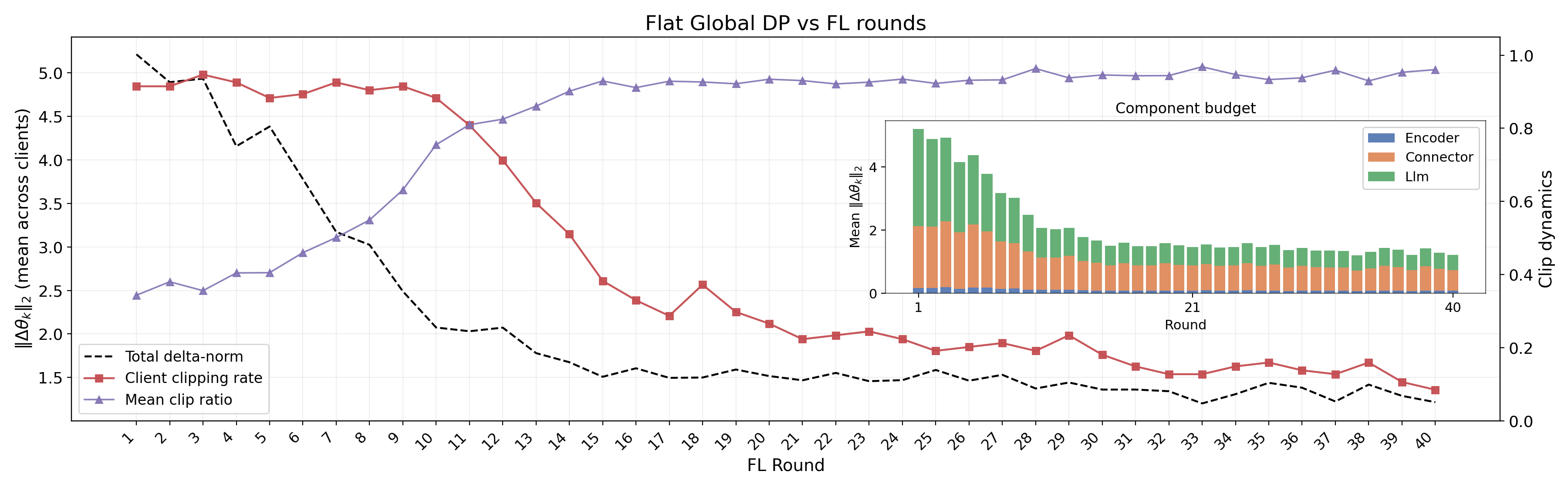}
  \caption{Flat DP with all unfrozen components (\textit{Flat-e} baseline, 40 rounds,
           $C{=}1.0$, averaging 94 clients/round). 
           Main panel (left axis, dashed): mean total joint update norm
           $\|\Delta\boldsymbol{\theta}^{(i)}\|_2$ across clients.
           Right axis: client clipping rate (red, fraction of clients
           whose update norm exceeds $C$) and mean clip ratio
           $\bar{r}^{(t)}\!=\!\frac{1}{n_t}\sum_i\min(1,C/\|\Delta\boldsymbol{\theta}^{(i)}\|_2)$
           (purple), i.e.\ the average fraction of each update that
           survives clipping before aggregation; $\bar{r}^{(t)}\!=\!1$
           means no client is clipped, while $\bar{r}^{(t)}\!\ll\!1$
           means updates are severely truncated.
           Inset: mean per-component norm budget per round
           (encoder blue, connector orange, LLM green).
           LLM and connector norms dominate throughout, sustaining a
           high global clip rate ($>$80\%) that suppresses the encoder
           signal to the noise floor.}
  \label{fig:flat-dp-clipping}
\end{figure*}

We can quantify this suppression via the per-component
\emph{signal-to-noise ratio} (SNR).  After aggregating $n$ clients,
the per-parameter signal of component $h$ is
$\|\widetilde{\Delta\boldsymbol{\theta}}_h\|_2/\sqrt{p_h}$, while the
per-parameter noise standard deviation from the Gaussian mechanism
\eqref{eq:gaussian_mech} is $\sigma C/n$.  Thus:
\begin{equation}
  \mathrm{SNR}_h
  = \frac{\displaystyle\left\|\widetilde{\Delta\boldsymbol{\theta}}_h\right\|_2
         /\,\sqrt{p_h}}
         {\sigma C / n}.
  \label{eq:snr}
\end{equation}
Thus the \emph{ratio} between two components depends
only on their clipped norms and parameter counts:
\begin{equation}
  \frac{\mathrm{SNR}_\mathcal{E}}{\mathrm{SNR}_\mathcal{L}}
  = \frac{\|\widetilde{\Delta\boldsymbol{\theta}}_\mathcal{E}\|_2\,/\,\sqrt{p_\mathcal{E}}}
         {\|\widetilde{\Delta\boldsymbol{\theta}}_\mathcal{L}\|_2\,/\,\sqrt{p_\mathcal{L}}}.
  \label{eq:snr-ratio}
\end{equation}
Under LLM-dominant global clipping, every client's update is scaled by
$r^{(i)} \approx C / \|\Delta\boldsymbol{\theta}_\mathcal{L}^{(i)}\|_2$,
so both components share the same $r^{(i)}$ and the ratio reduces to a norm-and-size comparison. With per-matrix norms differing one order of magnitude and similar per-matrix 
parameter counts (see Table~\ref{tab:params}), the per-layer encoder signal is ${\approx}10\times$ weaker than the LLM's.
Since the noise floor is identical for both pools, the encoder operates
at $\mathrm{SNR}_\mathcal{E} \approx 0.1\,\mathrm{SNR}_\mathcal{L}$
under global clipping. 
Fig.~\ref{fig:flat-dp-clipping} confirms that this structural imbalance persists dynamically under flat DP with all components unfrozen,
LLM+connector-dominated global norms keep the clipping rate high across all 40 rounds, leaving the encoder signal consistently noise-dominated during training.

\section{Evaluated DP Methods}
\label{sec:methods}

We perform FL-DP experiments using different strategies to allocate the $C$ budget across the layers of a Speech-LLM. All methods share common $C{=}1.0$ budget, $\sigma{=}0.1$, and server Gaussian noise $\mathcal{N}(0,(\sigma C/n)^2\mathbf{I})$. 
The methods in~\cite{DBLP:conf/nips/PelikanAFSTBL25} replace the single global clip in
\eqref{eq:global_clip} with per-layer $\ell$ budgets that partition $C^2$
proportionally to layer (LoRA rank matrix) size, $p_\ell$:
\begin{equation}
  C_\ell = C \cdot \sqrt{\frac{p_\ell}{\sum_k p_k}},
  \quad\text{so that}\quad \sqrt{\sum_\ell C_\ell^2} = C.
  \label{eq:pfl-dim}
\end{equation}
This \emph{Euclidean-tight} allocation~\cite{DBLP:conf/nips/PelikanAFSTBL25} was originally designed for
homogeneous single-component models with full-rank updates and is the
basis for all methods we study:

\begin{itemize}
    \item \textit{Flat}: refers to DP-FedAvg baseline~\cite{mcmahan2018learning}, i.e. a single norm $C$ is applied to the full concatenated update in Eq.~\ref{eq:flat_delta}, so no per-layer structure.
    \item \textit{PFL-Uniform}: we split uniformly the budget among LoRA matrices: $C_\ell = C/\sqrt{H}$, regardless of size. Where $H$ is the total number of matrices, i.e. adapted layers.
    \item \textit{PFL-Dim}: size-proportional allocation via Eq.~\eqref{eq:pfl-dim}; the ``dim-prop'' variant from~\cite{DBLP:conf/nips/PelikanAFSTBL25}, adapted for LoRA rank matrices.
    \item \textit{PFL-Unif+EMA} and \textit{PFL-Dim+EMA} (adaptive): we implemented original \emph{exponential moving average} (EMA) variants that dynamically reweight per-layer budgets from observed gradient norms.  At each FL round
$t$, a smoothed per-layer norm estimate is updated as
\begin{equation}
  \hat\nu_\ell^{(t)}
  = (1{-}\beta)\,\hat\nu_\ell^{(t-1)} + \beta\,\nu_\ell^{(t)},
  \label{eq:ema-update}
\end{equation}
where $\nu_\ell^{(t)}$ is the mean client delta-norm for layer $\ell$ in
round $t$, $\beta{=}0.2$ is the decay rate (higher $\beta$ = faster
adaptation).  Allocation weights and per-layer norms are then:
\begin{equation}
  w_\ell^{\text{target}} \propto
    \bigl(\hat\nu_\ell\bigr)^\gamma \cdot w_\ell^{\text{base}},
  \qquad C_\ell = C_t\sqrt{w_\ell},
  \label{eq:ema-weights}
\end{equation}
with $\gamma{=}0.75$ and $w_\ell^{\text{base}}$ equal to the PFL-Dim or PFL-Unif initialisation for each respective variant.  Weights are frozen at $w_\ell^{\text{base}}$ for the first three warmup rounds.
\item {\textit{$\alpha$-Split} (static): encoder and LLM$+$connector treated as independent pools with budgets
$C\sqrt{\alpha}$ and $C\sqrt{1{-}\alpha}$.  Fully described in following Section~\ref{sec:alphasplit}}.

\end{itemize}

The \textit{-e} suffix (e.g.\ \textit{PFL-Dim-e}) denotes the encoder-unfrozen variant; bare names denote encoder-frozen training.

\subsection{The $\alpha$-Split Design}
\label{sec:alphasplit}

\subsubsection{Formulation}
\label{sec:formulation}

Let $\alpha \in (0,1)$ be a \emph{component budget fraction} hyperparameter
controlling what share of $C^2$ is allocated to the encoder.  Encoder and LLM+connector parameters are clipped with \emph{independent} per-layer budgets:
\begin{align}
  C^{enc}_{\ell} = C \sqrt{\alpha} \sqrt{\frac{p_{\ell}}{\sum_{k \in \mathcal{E}} p_k}}, \quad \ell \in \mathcal{E}
\\[3pt]
C^{LLM}_{\ell} = C \sqrt{1-\alpha} \sqrt{\frac{p_{\ell}}{\sum_{k \in \mathcal{L}} p_k}}, \quad \ell \in \mathcal{L}
.
  \label{eq:alpha-llm}
\end{align}
Each pool is internally Euclidean-tight. Note that the connector is included in $\mathcal{L}$, because assigning it to $\mathcal{E}$ would incorrectly inflate the encoder budget. For the linear connector used in Whisper + \{TinyLlama,EuroLLM\} the two  matrices total ${\sim}4{,}096$
parameters and are indeed negligible within $\mathcal{L}$. 


\begin{table}[t]
\centering
\caption{Encoder and LLM (Whisper + TinyLlama) per-layer clip norms vs.\ $\alpha$.
         \textit{Flat-e} reference: $\bar{C}_\ell^{\text{LLM}} = 0.1474$.}
\label{tab:alpha}
\setlength{\tabcolsep}{4pt}
\begin{tabular}{lrrrl}
\toprule
$\alpha$ & $\bar{C}_\ell^{\text{enc}}$ & $\bar{C}_\ell^{\text{LLM}}$ & LLM / Flat-e & Note \\
\midrule
0.00 & 0.000 & 0.1503 & 1.020 & Enc.\ frozen \\
0.01 & 0.013 & 0.1495 & 1.014 & Min.\ enc.\ signal \\
\textbf{0.05} & \textbf{0.028} & \textbf{0.1465} & \textbf{0.994} & \textbf{$<$1\% LLM loss} \\
0.10 & 0.040 & 0.1426 & 0.967 & $<$5\% LLM loss \\
0.20 & 0.056 & 0.1345 & 0.912 & PFL-Dim+EMA-e steady-state \\
0.50 & 0.088 & 0.1063 & 0.721 & Half budget each \\
\bottomrule
\end{tabular}
\end{table}

\subsubsection{Sensitivity and Privacy Preservation}
\label{sec:sensitivity}

Let $c_{\mathcal{E}} = \|\text{clipped}_\mathcal{E}\|_2$ and $c_{\mathcal{L}} = \|\text{clipped}_\mathcal{L}\|_2$. 
Then the $\ell_2$ sensitivity is:
\begin{equation}
  \Delta = \sqrt{c_{\mathcal{E}}^2 + c_{\mathcal{L}}^2}
           \leq \sqrt{\alpha C^2 + (1-\alpha) C^2} = C.
  \label{eq:sensitivity}
\end{equation}
The sensitivity is unchanged at $C$. The server Gaussian noise $\mathcal{N}(0,(\sigma C/n)^2\mathbf{I})$ requires no modification, and the Rényi DP accountant, $\varepsilon$, and $\delta$ are all identical
to \textit{Flat-DP}. Thus $\alpha$ is a pure utility-vs-component-privacy parameter.

\subsubsection{Choosing $\alpha$}
\label{sec:alpha-choice}

Table~\ref{tab:alpha} shows the LLM per-layer budget as a function of $\alpha$.  We select $\alpha = 0.05$ because (i) the LLM retains 99.4\% of its \textit{Flat-e} budget with negligible regression risk; (ii) the encoder clip norm ($0.028$) is ${\approx}2\times$ the observed encoder delta-norm ($0.013$), providing sufficient gradient signal without near-100\% clipping. Note that both criteria depend solely on parameter counts and the encoder/LLM update-norm ratio, which can be measured on any public corpus before deployment, making $\alpha$ selection fully privacy-compatible.


\begin{figure}[t!]
  \centering
  \includegraphics[width=\linewidth]{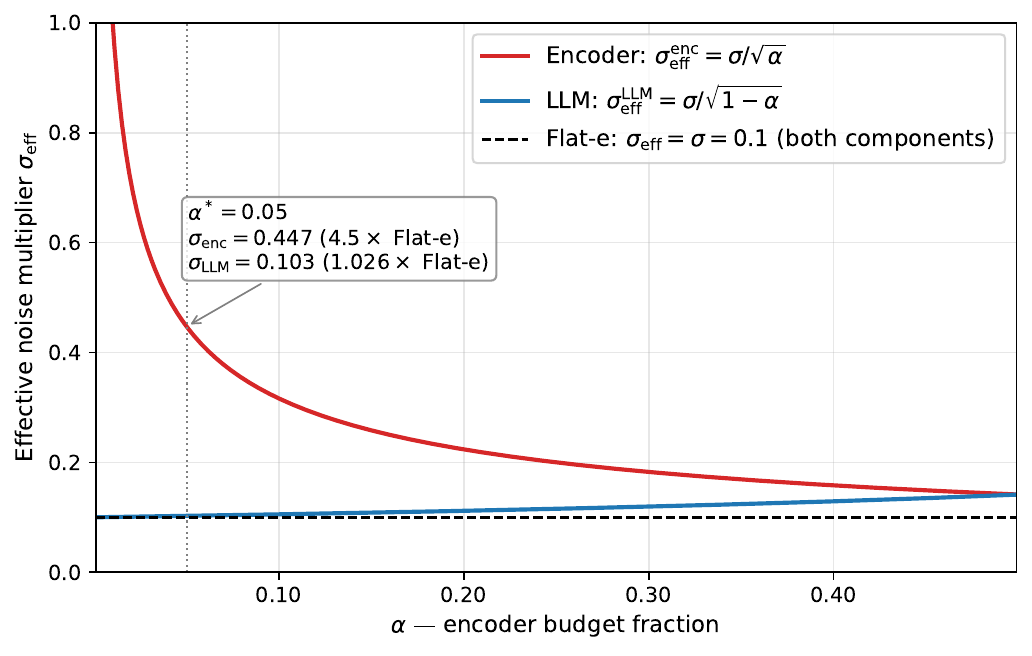}
  \caption{Effective per-component noise multiplier vs.\ encoder budget
           fraction $\alpha$.  Solid lines: \textit{$\alpha$-Split-e} (encoder
           red, LLM blue).  Dashed: \textit{Flat-e} reference
           ($\sigma_\mathrm{eff}{=}\sigma$ for both).  At $\alpha^*{=}0.05$
           (vertical dotted line) the encoder receives $4.47\times$
           tighter per-component privacy at only $+2.6\%$ LLM overhead,
           with no change to the joint $(\varepsilon,\delta)$ guarantee.}
  \label{fig:sigma-eff}
\end{figure}

\subsubsection{Per-Component Privacy Advantage}
\label{sec:privacy}
Note that \textit{Flat}-DP should achieve the utility ceiling (lower WER for ASR task) for any per-layer strategy at the same $(\varepsilon,\delta)$. Any per-layer method at the same global sensitivity $C$ calibrates noise to $C$, so $\min(\sigma_\mathrm{eff}^\mathcal{E},\,\sigma_\mathrm{eff}^\mathcal{L}) \geq \sigma$. The genuine advantage of $\alpha$-split is \emph{asymmetric per-component privacy}.  Because the encoder pool is bounded at $C\sqrt{\alpha}$, its effective noise multiplier is $\sigma_\mathrm{eff}^\mathcal{E} = \frac{\sigma}{\sqrt{\alpha}}
 = 0.447$, which is $4.47\times$ that of \textit{Flat-e}, while the LLM pays only $\sigma_\mathrm{eff}^\mathcal{L} = \frac{\sigma}{\sqrt{1-\alpha}}
 = 0.103$, which is only $+2.6\%$ vs.\ \textit{Flat-e}. Fig.~\ref{fig:sigma-eff} visualises $\sigma_\mathrm{eff}$ as a function of $\alpha$, confirming the steep encoder privacy gain and flat LLM overhead for small $\alpha$.
Since the encoder pool is bounded at $C\sqrt{\alpha}$, an adversary performing gradient inversion learns $4.47\times$ less about acoustic encoder updates, directly guarding speaker biometric attributes (accent, prosody, voice identity).
Note that what \textit{$\alpha$-split} changes is the \emph{per-component} interpretation: standard global clipping mixes acoustic and linguistic gradients into a single pool, whereas $\alpha$-split structurally isolates the encoder's biometric representations, providing a privacy guarantee that per-layer single-pool methods cannot achieve regardless of their budget allocation.

\begin{figure}[t]
  \centering
  \includegraphics[width=\linewidth]{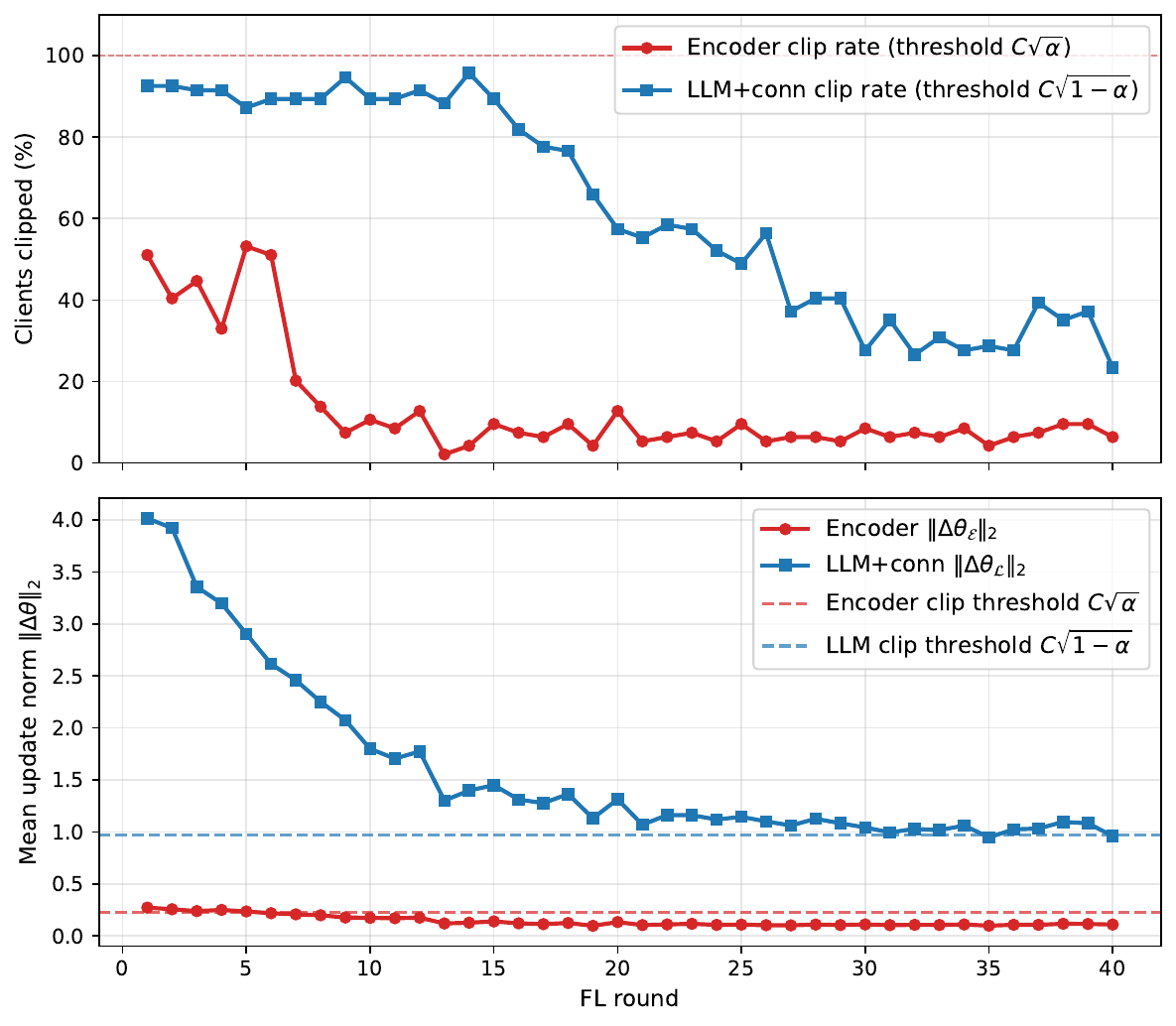}
  \caption{\textit{Top:} Per-component client clipping rate per FL round
           under \textit{$\alpha$-Split-e} ($\alpha{=}0.05$, $C{=}1.0$, 94 clients/round).
           Red: encoder pool ($C\sqrt{\alpha}{=}0.224$);
           blue: LLM+connector pool ($C\sqrt{1{-}\alpha}{=}0.975$).
           \textit{Bottom:} Mean update norm $\|\Delta\boldsymbol{\theta}\|_2$
           per component---encoder $\|\Delta\boldsymbol{\theta}_{\mathcal{E}}\|_2$
           (red) and LLM+connector $\|\Delta\boldsymbol{\theta}_{\mathcal{L}}\|_2$
           (blue)---with dashed lines at the respective pool clip thresholds.}
  \label{fig:clip-rate}
\end{figure}

\subsubsection{Training Dynamics and Clipping Rate Analysis}
\label{sec:clip-rate}

Fig.~\ref{fig:clip-rate} shows the per-component clipping rate vs.\ FL round for the \textit{$\alpha$-Split-e} run ($\alpha{=}0.05$, $C{=}1.0$, averaging 94 clients/round). Two distinct regimes emerge: the encoder clip rate starts at ${\approx}51\%$
(rounds 1--6) and quickly falls to $5$--$14\%$ once its norm stabilises well below the $0.224$ threshold; the LLM+connector starts at ${\approx}93\%$ clip rate (norm ${\approx}4.0$) and converges gradually to $25$--$40\%$ by round~40. The low encoder clipping rate after round 8 confirms that \textit{$\alpha$-Split} is structurally non-destructive. Because DP guarantees depend on bounding worst-case sensitivity rather than active clip frequency, this mechanism enforces the biometric privacy bound while keeping clipping bias minimal once training stabilises.

\section{Experimental Results}
\label{sec:experiments}

Table~\ref{tab:wer} compares the utility of all methods on the MLS test split. Statistical significance of WER differences is assessed using speaker-level percentile bootstrap test speakers ($B{=}10{,}000$, 95\% CI), with all utterances from each sampled speaker included to preserve within-speaker correlations~\cite{bisani2004bootstrap}.

\subsection{Encoder-Frozen Variants}
\label{sec:frozen}

When the encoder is frozen, cross-component coupling is absent and results are consistent across all three architectures. \textit{PFL-Dim+EMA} achieves the best DP result in every group (W+TinyLlama: $0.1657$; W+EuroLLM: $0.1362$; Voxtral: $0.1450$), outperforming \textit{Flat} by $0.0028$--$0.0082$. \textit{PFL-Uniform} degrades severely in all frozen groups ($0.4643$ for W+TinyLlama,
$0.3500$ for W+EuroLLM), confirming equal-budget-per-layer is harmful regardless of architecture. Notably, \textit{$\alpha$-Split} slightly \emph{underperforms} \textit{Flat} for EuroLLM frozen ($+0.0128$ vs.\ $+0.0114$): with the encoder inactive the entire budget falls on LLM layers under any method, so the two-pool design yields no structural advantage.

\subsection{Encoder-Unfrozen Variants and Cross-Component Collapse}
\label{sec:unfrozen}

Activating the encoder exposes the cross-component budget collapse across all
single-pool methods.
On Whisper+TinyLlama, \textit{$\alpha$-Split-e} remains the \emph{only}
per-layer method that avoids catastrophic degradation.
\textit{PFL-Uniform-e}/\textit{PFL-Unif+EMA-e} collapse ($0.9944^\dagger$/$0.8243^\dagger$);
\textit{PFL-Dim-e} and \textit{PFL-Dim+EMA-e} regress $+63\%$ and $+30\%$
vs.\ \textit{Flat-e}. The \textit{PFL-Uniform-e}$\to$\textit{PFL-Unif+EMA-e}
comparison suggests that EMA cannot rescue a structurally broken base allocation.
\textit{$\alpha$-Split-e} stays within $0.003$ absolute of the \textit{Flat-e} ($0.1959$),
the utility ceiling for any per-layer method at the same $(\varepsilon,\delta)$
(Section~\ref{sec:privacy}). 
Repeating \textit{Flat-e}, \textit{PFL-Dim+EMA-e}, and \textit{$\alpha$-Split-e} with two additional random seeds yields std\,$\leq 0.02$ for all three, confirming the rankings are stable across training runs.

On Whisper+EuroLLM the same replicates on a stronger multilingual backbone, confirming that cross-component degradation is architectural. \textit{PFL-Dim-e} ($0.1975$) and \textit{PFL-Dim+EMA-e} ($0.1845$) regress
vs.\ \textit{Flat-e} ($0.1457$); \textit{PFL-Unif+EMA-e} collapses to $0.4769$. \textit{$\alpha$-Split-e} ($0.1600$) is again the only per-layer method avoiding
regression, reducing the gap to \textit{Flat-e} from $+0.0388$ to $+0.0143$. Both \textit{Flat-e} ($-0.0300$) and \textit{$\alpha$-Split-e} ($-0.0157$) \emph{outperform} the no-DP FedAvg baseline ($0.1757$), consistent with gradient clipping acting as implicit regularisation under the unstable default encoder lr.

\begin{table}[t]
\centering
\caption{WER on MLS \texttt{test} split (19{,}492~samples, $C{=}1.0$, $\sigma{=}0.1$, 40~FL rounds). Gap = WER$-$FedAvg~(no~DP) per group. $\dagger$: collapsed ($\mathrm{WER}{>}0.5$). $*$: mean over 3 seeds; std\,$\leq\!0.02$. \textbf{Bold}: best DP result per group/column; {---}: not evaluated.}

\label{tab:wer}
\small
\setlength{\tabcolsep}{5pt}
\begin{tabular}{lrrrr}
\toprule
& \multicolumn{2}{c}{\textbf{Enc.\ frozen}} & \multicolumn{2}{c}{\textbf{Enc.\ unfrozen}} \\
\cmidrule(lr){2-3}\cmidrule(lr){4-5}
\textbf{Method} & \textbf{WER} & \textbf{Gap} & \textbf{WER} & \textbf{Gap} \\
\midrule
\multicolumn{5}{l}{\textit{Whisper+TinyLlama-1.1B}} \\
FedAvg (no DP)  & 0.1415 & ---       & 0.1842              & ---                 \\
Flat            & 0.1685 & $+0.0270$ & $0.1959^{*}$              & $+0.0117$           \\
PFL-Uniform     & 0.4643 & $+0.3228$ & $0.9944^\dagger$    & $+0.8102^\dagger$   \\
PFL-Dim         & 0.1711 & $+0.0296$ & 0.3051              & $+0.1209$           \\
PFL-Dim+EMA     & \textbf{0.1657} & $+0.0242$ & $0.2599^{*}$   & $+0.0757$           \\
PFL-Unif+EMA    & 0.2360 & $+0.0945$ & $0.8243^\dagger$    & $+0.6401^\dagger$   \\
$\alpha$-Split  & 0.1677 & $+0.0262$ & $\mathbf{0.1986}^{*}$     & $+0.0144$           \\
\midrule
\multicolumn{5}{l}{\textit{Whisper+EuroLLM-1.7B-Instruct}} \\
FedAvg (no DP)  & 0.1330 & ---       & 0.1757              & ---                 \\
Flat            & 0.1444 & $+0.0114$ & 0.1457              & $-0.0300$           \\
PFL-Uniform     & 0.3500 & $+0.2170$ & ---                 & ---                 \\
PFL-Dim         & 0.1381 & $+0.0051$ & 0.1975              & $+0.0218$           \\
PFL-Dim+EMA     & \textbf{0.1362} & $+0.0032$ & 0.1845   & $+0.0088$           \\
PFL-Unif+EMA    & 0.1780 & $+0.0450$ & 0.4769              & $+0.3012$           \\
$\alpha$-Split  & 0.1458 & $+0.0128$ & \textbf{0.1600}     & $-0.0157$           \\
\midrule
\multicolumn{5}{l}{\textit{Voxtral-Mini-3B}} \\
FedAvg (no DP)  & 0.1442 & ---       & 0.1362              & ---                 \\
Flat            & 0.1478 & $+0.0036$ & 0.1467              & $+0.0105$           \\
PFL-Dim+EMA     & \textbf{0.1450} & $+0.0008$ & \textbf{0.1413} & $+0.0051$    \\
$\alpha$-Split  & ---    & ---       & 0.1495              & $+0.0133$           \\
\bottomrule
\end{tabular}
\end{table}
On Voxtral-Mini-3B the regime differs: a dedicated no-DP profiling run confirms
a steady-state LLM/encoder norm ratio of only ${\approx}1.7{\times}$
(vs.\ ${\approx}12{\times}$ for Whisper+TinyLlama).
At this milder imbalance, \textit{PFL-Dim+EMA-e} does \emph{not} collapse
($0.1413$, $+0.0051$); EMA adaptation suffices without structural pool separation.
\textit{$\alpha$-Split-e} ($0.1495$) is the \emph{worst} DP method: the fixed
$\alpha{=}0.05$ over-clips the encoder (clip/norm ${\approx}0.04$) whose gradient
magnitude is comparable to the LLM at this ratio.
Together, the three architectures delineate the operating regime: $\alpha{=}0.05$
is optimal when the encoder/LLM norm ratio is ${\geq}12{\times}$;
\textit{PFL-Dim+EMA-e} is the safer default when the ratio is unknown or
substantially below $12{\times}$.

\section{Discussion}
\label{sec:discussion}

Although $\alpha = 0.05$ is well-calibrated for the extreme ${\approx}12{\times}$ norm imbalance in Whisper + TinyLlama architecture, it should not be applied blindly. In practice, $\alpha$ can be selected dynamically by observing component delta-norms during a few non-DP warmup rounds, as shown in Table~\ref{tab:alpha}. Architectures with milder imbalances (e.g., Voxtral at ${\approx}1.7{\times}$) do not require a structural pool split, and adaptive single-pool methods like \textit{PFL-Dim+EMA} suffice. Blindly imposing $\alpha = 0.05$ in such regimes over-clips the encoder, as confirmed by our Voxtral results in Table~\ref{tab:wer}.

While unfreezing the encoder under non-private FedAvg degrades performance due to default learning rate sensitivity in Whisper + TinyLlama (yet improves Voxtral-Mini-3B), DP clipping actually outperforms this baseline on EuroLLM. This might appear to indicate a poorly tuned baseline, but controlled learning-rate sweep shows that optimising the encoder lr (${\times}0.02$) brings no-DP FedAvg to WER~$0.1196$, far below every DP result, see Appendix for further details. The DP-over-FedAvg result at default lr reflects that DP sensitivity bounds act as an implicit regularizer, dampening divergent encoder updates under unstable optimisation scales~\cite{bu2022automatic}. The Table~\ref{tab:wer} comparison is internally consistent: all DP methods use the same default lr as the FedAvg baseline, so the relative DP--FedAvg gap is meaningful; only the absolute WER level would shift under a tuned lr. Crucially, encoder learning rate tuning and $\alpha$-split are complementary, not interchangeable. Reducing the encoder learning rate (e.g., by $50{\times}$) shrinks the cross-component norm gap without altering the DP mechanism, but it lowers the encoder's gradient signal-to-noise ratio under DP noise. By contrast, our proposed $\alpha$-split formulation preserves the encoder's full learning rate while granting a $4.47{\times}$ tighter component-specific privacy protection, structurally resolving the bottleneck of excessive gradient truncation.

\section{Conclusion}
\label{sec:conclusion}

We identified \emph{cross-component budget collapse} as a  failure of single-pool per-layer DP clipping in Speech-LLMs, where encoder updates dilute the LLM clipping budget and cause severe utility regressions. Our proposed \textit{$\alpha$-Split-e} design resolves this pathology by decoupling the encoder and LLM updates into independent pools. At an architecture-calibrated $\alpha = 0.05$, \textit{$\alpha$-Split-e} achieves utility close to both flat DP and non-DP baselines on Whisper+TinyLlama, while providing $4.47\times$ tighter biometric privacy protection for the speech encoder. On the Whisper+EuroLLM architecture, \textit{$\alpha$-Split-e} yields the best per-layer DP results, outperforming our adaptive \textit{PFL-Dim+EMA-e} baseline by $2.45\%$ absolute—confirming that cross-component budget collapse is structural and independent of the LLM backbone. Conversely, on Voxtral-Mini-3B, which exhibits a mild gradient-norm imbalance (${\approx}1.7{\times}$), our adaptive \textit{PFL-Dim+EMA-e} formulation achieves the best overall DP utility. Our findings establish clear deployment boundaries: a structural \textit{$\alpha$-split} is optimal for extreme component imbalances ($\ge 12{\times}$), whereas the dynamic \textit{PFL-Dim+EMA-e} serves as the more robust default for balanced architectures.

\section*{AI-Generated Content Disclosure}
We used a generative AI to assist in paraphrasing, improving clarity and grammar in parts of the manuscript and coding assistance. All generated content was reviewed and validated by the authors.

\section*{Acknowledgment}
This work has received funding from the European Union's Horizon Europe research and innovation programme under the project ELOQUENCE (Grant Agreement No. 101135916). This work was supported by computational resources from the EuroHPC Joint Undertaking under the EuroHPC AI Factory grant EHPC-AIF-2026LS01-004.

\bibliographystyle{IEEEtran}
\bibliography{references}

\appendices
\section{EuroLLM Encoder Learning-Rate Sensitivity}
\label{sec:lrsweep}

The EuroLLM unfrozen results in Table~\ref{tab:wer} use a default encoder
learning rate equal to the LLM's (multiplier $\times 1.0$).  Because
EuroLLM-1.7B produces stronger LLM gradient signals than TinyLlama-1.1B,
the encoder is dominated in training unless its learning rate (lr) is reduced. Table~\ref{tab:lrsweep} shows WER as a function of the encoder learning rate (enc-lr) multiplier for FedAvg (no DP, client local epochs $E=10$, over 40 FL rounds). The best non-DP result, enc-lr $\times 0.02$ / le5 (WER $0.1196$), closely matches the frozen EuroLLM baseline ($0.1330$) and substantially improves over the default
lr baseline ($0.1757$, Table~\ref{tab:lrsweep}).
\begin{table}[h!]
\centering
\caption{Non-DP FedAvg WER vs. encoder learning rate multiplier, Whisper+EuroLLM unfrozen (MLS test partition, 40 rounds).}
\label{tab:lrsweep}
\setlength{\tabcolsep}{4pt}
\begin{tabular}{lrr}
\toprule
Enc.\ LR mult.\ & Local ep.\ & WER \\
\midrule
$\times 0.02$           & 5  & \textbf{0.1196} \\
$\times 0.02$           & 10 & 0.1264 \\
$\times 0.05$           & 5  & 0.1325 \\
$\times 0.10$           & 5  & 0.1343 \\
$\times 0.20$           & 5  & 0.1598 \\
$\times 0.02$ (enc+llm) & 10 & 0.1393 \\
$\times 1.00$ (default) & 10 & 0.1757 \\
$\times 1.00$ (default) & 5  & 0.1861 \\
\bottomrule
\end{tabular}
\end{table}

Table~\ref{tab:dp_lr002} reports DP
results at enc-lr $\times 0.02$ (encoder only), using the matched no-DP baseline of
WER $0.1264$ (enc-lr $\times 0.02$, E=10).  PFL-Dim+EMA-e improves dramatically from
$0.1845$ (default lr, Table~\ref{tab:wer}) to $\mathbf{0.1440}$ ($+0.0176$), now
\emph{outperforming} both \textit{$\alpha$-Split-e} ($0.1511$, $+0.0247$) and
\textit{Flat-e} ($0.1525$, $+0.0261$).

\begin{table}[h!]
\centering
\caption{DP WER at encoder learning rate $\times 0.02$ (encoder only), Whisper+EuroLLM unfrozen
         ($C{=}1.0$, $\sigma{=}0.1$, E=10).
         Gap vs.\ matched no-DP baseline (enc-lr $\times 0.02$, le10, WER $0.1264$).
         \textbf{Bold}: best DP result.}
\label{tab:dp_lr002}
\setlength{\tabcolsep}{4pt}
\begin{tabular}{lrr}
\toprule
\textbf{Method} & \textbf{WER} & \textbf{Gap} \\
\midrule
FedAvg (no DP)   & 0.1264 & --- \\
Flat-e           & 0.1525 & $+0.0261$ \\
\textbf{PFL-Dim+EMA-e} & \textbf{0.1440} & $+0.0176$ \\
$\alpha$-Split-e & 0.1511 & $+0.0247$ \\
\bottomrule
\end{tabular}
\end{table}

This reversal result directly confirms the cross-component collapse diagnosis. At the default encoder lr ($\times 1.0$), encoder update norms are $\sim$10$\times$ larger than individual LLM-layer norms; the 64 encoder matrices monopolise clip-budget slots in the single shared pool, leaving the LLM under-served.  Reducing the encoder lr by $\times 50$ shrinks encoder update norms proportionally, eliminating the inter-component norm
imbalance \emph{without any change to the DP mechanism}.  With the imbalance
removed, \textit{PFL-Dim+EMA-e}'s global adaptive allocation can concentrate the full
budget~$C$ on the LLM layers where gradients are large, recovering near-flat utility.  By contrast, \textit{$\alpha$-Split-e} enforces a fixed $5\%$/$95\%$ (tuned on Whisper  + TinyLlama architecture) encoder/LLM pool split regardless of actual norms; with encoder updates already attenuated by the reduced lr, the encoder pool is structurally under-utilised and the LLM pool is marginally tighter than a globally adaptive method would choose---giving \textit{PFL-Dim+EMA-e} a small but consistent edge.

Importantly, lr tuning and $\alpha$-split are complementary, not
interchangeable: the $\times 50$ lr reduction attenuates the encoder's gradient signal (lower signal-to-noise ratio under DP noise), whereas $\alpha$-split preserves the encoder's full lr while granting it $4\times$ tighter
$(\varepsilon,\delta)$-DP protection.  The enc-lr $\times 0.02$ result therefore serves as a controlled ablation: it proves that the collapse observed at default lr is driven by the encoder/LLM gradient-norm imbalance.
\section{MLS Corpus Granular Statistics and Speaker Overlap Analysis}
\label{appendix:mls_statistics}

This appendix provides the granular structural details of the Multilingual LibriSpeech (MLS) speaker-based client partition ($K=316$ clients) to complement the high-level dataset overview presented in Section II-C.

\subsection{Client Volume and Linguistic Imbalances}
The stratified speaker partition induces an extreme, multi-dimensional data imbalance across both languages and individual clients. While English clients dominate the network numerically, they represent a small fraction of the total training hours. Conversely, Continental European languages are represented by a few highly active clients with massive local datasets. 

Specifically, the partition exhibits the following structural characteristics:
\begin{itemize}
    \item \textbf{High-Client, Low-Volume Regimes (English):} English represents $81.0\%$ of the total client population ($256$ out of $316$ clients) but accounts for only $15.4\%$ ($105.3$~hours) of the total training data. This yields an average training volume of only $0.41$~hours ($24.7$~minutes) per English client.
    \item \textbf{Low-Client, High-Volume Regimes (Polish):} Polish represents exactly $0.31\%$ of the client population ($1$ client) but holds $3.7\%$ of the total training data ($25.7$~hours). The single Polish client has over $62\times$ more training audio than the average English client.
    \item \textbf{Highly Symmetric Clusters (German/Spanish):} German and Spanish exhibit a moderate balance between client counts and data volume, with Spanish clients averaging $9.31$~hours and German clients averaging $8.47$~hours.
\end{itemize}

This severe volume disparity (ranging from $24$ minutes to over $25$~hours per client) provides a challenging non-IID optimization landscape. Standard FedAvg updates are weighted by the local client sample count $n_k$, meaning a tiny fraction of highly active European clients can disproportionately influence global gradient updates during a given round. Table~\ref{tab:mls_statistics_appendix} outlines the complete language-by-language breakdown of training hours, client counts, and average local data volumes.

\begin{table}[htbp]
\centering
\caption{Granular breakdown of the MLS training pool under the speaker-based partition ($K = 316$ clients).}
\label{tab:mls_statistics_appendix}
\begin{tabular}{lrrr}
\toprule
\textbf{Language} & \textbf{Training Hours} & \textbf{Clients ($K$)} & \textbf{Avg. Hours/Client} \\
\midrule
French     & 251.6 & 15  & 16.77 \\
German     & 160.9 & 19  & 8.47  \\
English    & 105.3 & 256 & 0.41  \\
Spanish    & 83.8  & 9   & 9.31  \\
Italian    & 27.3  & 7   & 3.90  \\
Polish     & 25.7  & 1   & 25.70 \\
Portuguese & 18.5  & 5   & 3.70  \\
Dutch      & 12.7  & 4   & 3.18  \\
\midrule
\textbf{Total} & \textbf{685.7} & \textbf{316} & \textbf{2.17} \\
\bottomrule
\end{tabular}
\end{table}

\subsection{Speaker Overlap Mechanics in audiobook Corpora}
Because the MLS corpus is derived from public-domain LibriVox audiobooks, the training, validation, and evaluation splits are subject to the inherent constraints of audiobook recording structures. In LibriVox, a single volunteer reader (speaker) frequently contributes to multiple books, or records independent chapters across different volumes. 

To maximize the acoustic diversity of the dataset, established central ASR benchmarks split these recorded chapters across training, validation, and test sets. When simulating a speaker-based cross-device FL network, where each unique speaker is mapped to a single client node, this underlying splits design introduces a $2.5\%$ speaker overlap. Specifically, $8$ out of the $316$ clients ($2.8\%$ of total training samples) represent speakers who also appear in the test split.

We explicitly maintain this standard partition for three critical methodological reasons:
\begin{enumerate}
    \item \textbf{Generalization vs. Personalization:} It allows us to evaluate the model's generalized performance on entirely unseen speakers, while simultaneously observing how well the model adapts to unseen utterances from speakers who were present in the training set.
    \item \textbf{Disjoint Utterances:} No individual audio sample (utterance) is shared between splits. The validation and test utterances are entirely disjoint from the training data, preventing direct memorization or data leakage.
    \item \textbf{Benchmark Alignment:} Preserving this speaker allocation ensures our FL results are directly comparable to existing centralized and federated benchmarks in the ASR literature, maintaining empirical continuity.
\end{enumerate}

\end{document}